# Using Deep Learning to Predict Plant Growth and Yield in Greenhouse Environments

Bashar Alhnaity[1], Simon Pearson[2], Georgios Leontidis[1] and Stefanos Kollias[1]

[1]School of Computer Science, University of Lincoln, Lincoln, LN6 7TS, UK
[2]Lincoln Institute for Agri-Food Technology, University of Lincoln, Lincoln, LN6 7TS, UK.

**Abstract**
**Effective plant growth and yield prediction is an essential task for greenhouse growers and for agriculture in general. Developing models which can effectively model growth and yield can help growers improve the environmental control for better production, match supply and market demand and lower costs. Recent developments in Machine Learning (ML) and, in particular, Deep Learning (DL) can provide powerful new analytical tools. The proposed study utilises ML and DL techniques to predict yield and plant growth variation across two different scenarios, tomato yield forecasting and Ficus benjamina stem growth, in controlled greenhouse environments. We deploy a new deep recurrent neural network (RNN), using the Long Short-Term Memory (LSTM) neuron model, in the prediction formulations. Both the former yield, growth and stem diameter values, as well as the microclimate conditions, are used by the RNN architecture to model the targeted growth parameters. A comparative study is presented, using ML methods, such as support vector regression and random forest regression, utilising the mean square error criterion, in order to evaluate the performance achieved by the different methods. Very promising results, based on data that have been obtained from two greenhouses, in Belgium and the UK, in the framework of the EU Interreg SMARTGREEN project (2017-2021), are presented.**

**INTRODUCTION**

As with many bio-systems, plant growth is a highly complex and dynamic environmentally linked system. Therefore, growth and yield modeling is a significant scientific challenge . Modeling approaches vary in a number of aspects (including, scale of interest, level of description, integration of environmental stress, etc.). According to (Todorovski and Dzeroski, 2006; Atanasova et al., 2008) two basic modeling approaches are possible, namely, "knowledge-driven" or "data-driven" modeling. The knowledge driven approach relies mainly on existing domain knowledge. In contrast, a data-driven modeling approach is capable of formulating a model solely from gathered data without necessarily using domain knowledge.

Data driven models (DDM) include classical Machine Learning techniques, artifical neural networks (Daniel et al., 2008), support vector machines (Pouteau et al., 2012), and generalized linear models. Those methods have many desirable characteristics, such as imposing fewer restrictions, or assumptions, the ability to approximate nonlinear functions, strong predictive abilities, and the flexibility to adapt to inputs of a multivariate system (Buhmann, 2003). According to Singh et al., 2016 and reviewed by Liakos et al., 2018 Machine Learning (ML), linear polarizations, wavelet-based filtering, vegetation indices (NDVI) and regression analysis are the most popular techniques used for analyzing agricultural data. However and besides the aforementioned techniques, a new methodology which is recently gaining momentum is deep learning (DL)(Goodfellow et al., 2016). DL belongs to the machine learning computational field and is similar to ANN. However, DL is about "deeper" neural networks that provide a hierarchical representation of the data by means of various operations. This allows larger learning capabilities, and thus higher performance and precision. A strong advantage of DL is feature learning, i.e., automatic feature extraction from raw data, with features from higher levels of the hierarchy being formed by composition of lower level features (Goodfellow et al., 2016). DL can solve more complex problems particularly well, because of the more complex related models (Pan and Yang, 2010). These complex models employed in DL can increase classification accuracy and reduce error in regression problems, provided there are adequately large data-sets available describing the problem.

Gonzalez-Sanchez et al.( 2019) presented a comparative study of ANN, SVR, M5-prime, KNN ML techniques and Multiple Linear Regression for crop yield prediction in ten crop datasets. In their study, Root Mean Square Error (RMS), Root Relative Square Error (RRSE), Normalized Mean Absolute Error (MAE) and Correlation Factor (R) were used as accuracy metrics to validate the models. Results showed that M5-Prime achieved the lowest errors across the produced crop yield models. The results of that study ranked the techniques from the best to the worst, according to RMSE, RRSE, R, and MAE resulting, in the following order: M5-Prime, kNN, SVR, ANN and MLR. Another study by (Nair and Yang-Won, 2016) applied four ML techniques, SVM, Random Forest (RF), Extremely Randomized Trees (ERT) and Deep Learning (DL) to estimate corn yield in Iowa State. Comparisons of the validation statistics showed that DL provided more stable results, overcoming the overfitting problem.

Stem diameter is considered as one of the important parameters describing the growth of plants during vegetative growth stage. Also, the variation of stem diameter has widely been used to derive proxies for plant water status and, is therefore applied in optimisation strategies for plant-based irrigation scheduling in a wide range of species. Plant stem diameter variation (SDV) refers to plant stem periodic shrinkage and recovery movement during the day and night, and this periodic variation is related to plant water content and can be used as an indicator of the plant water content change. During active vegetative growth and development, crop plants rely on the carbohydrate gained from photosynthesis and the translocation of photo-assimilates from the site of synthesis to sink organs (Yu et al., 2015). The fundamentals of stem diameter variations have been well documented in a substantial amount of literature (Vandegehuchet et al., 2014). It has been documented that SDV is sensitive to water and nutrient conditions and is closely related to the responses of crop plants to the changes of environmental conditions (Kanai et al., 2008). The stem diameter is an important parameter describing the growth of crop plants under abiotic stress during vegetative growth stage. Therefore, it is important to generate stem diameter growth models able to predict the response of SDV to environmental changes and plant growth under different conditions. Many studies emphasize the need to critically review and improve SDV models for assessment of environmental impact on crop growth (Hinckley and Bruckerhoff, 2011). SDV daily models have been developed to accurately predict inter-annual variation in annual growth in balsam fir (Abies balsamea L) (Duchesene and Houle, 2011). Inclusion of daily data in growth-climate models can improve predictions of the potential growth response to climate by identifying particular climatic events that escape to a classical dendroclimatic approach (Duchesene and Houle, 2011). However, models for predicting SDV and plant growth using environmental variables have so far remained limited.

Tomato crop growing in greenhouse environment is considered as a dynamic and complex system, with few models having been studied for it up to now. In the literature TOMGRO and TOMSIM (Jones et al., 1999), (Heuvelink, 1996) are considered as the main applicable dynamic growth models. Those models are dependent on physiological processes, and they represent biomass partitioning, crop growth, and yield as a function of several climate and physiological parameters. However, due to their limited application to practical settings, their complexity, the difficulty in estimating initial parameter values and the need for calibration and validation in every new environment, growers uptake has been limited.

The Tompousse model was developed by (Abreu et al., 2000) to predict tomato yield in terms of the weight of harvested fruits. The model was developed by examining the relationship between environmental parameters in a heated greenhouses in the Southern part of France. A linear relationship between flowering rate and fruit growth was the basic assumption used in this model. However, the model performance was poor when tested in unheated plastic greenhouses in Portugal. Another tomato yield model was proposed by Adams (Adams, 2002), based on a form of graphical simulation tool. The main objective of the model was to represent weekly fluctuations of greenhouse tomato yield in terms of fruit size and harvest rate. Hourly climate data were used to estimate the rate of growth of leaf truss and the flower production. Yield seasonal fluctuations were generally infuenced by periodic variations of solar radiation and air temperature. According to (Qaddoum et al., 2013), there is a large number of tools that can help farmers in making decisions. These can provide yield rate prediction, suggest climate control strategies, synchronise crop production with market demands.

A deep learning model is proposed in this paper, which is trained with environmental ($CO_2$, humidity, radiation, outside temperature, inside temperature), as well as, actual yield and stem diameter variation measurements and has the ability to produce accurate prediction of either ficus stem diameter, or tomato yield problems. The rest of this paper is organized as follows. In Section 2, we briefly describe the proposed approach and the utilised datasets. Section 3 presents the obtained results, and Section 4 presents conclusions and future work.

**MATERIALS AND METHODS**

**Conventional Machine Learning**

One of the main advantages of Machine learning (ML) techniques is that they are capable of autonomously solving large non-linear problems using datasets from multiple sources. ML enables better decision making and informed action in real-world scenarios without (or with minimal) human intervention. It provides a powerful and flexible framework for data-driven decision making that can be widely used, also being highly applicable in agricultural applications. In recent years different ML techniques have been implemented to achieve accurate plant growth, yield and production prediction for different crops. As already mentioned, the most successful techniques are Artificial Neural Networks, Support Vector Regression (SVR), M5-prime Regression Trees, Random Forests (RF), and K-Nearest Neighbors (Chlingaryan et al., 2018). In this paper, SVR and RF models as baseline models to predict plant yield and growth are used.

**Support vector regression (SVR)**

Support vector regression (SVR) arises from a nonlinear generalization of the Generalized Portrait algorithm developed by Vapnik (Cortes and Vapnik, 1995). It projects the input data into a higher dimensional space using a kernel function and separates different classes of data using a hyperplane. The trade-off between margin and errors is controlled by the regularization parameter c. SVR with radial basis kernel functions($SVR_{rbf}$) uses $K(x_i, x_j) = \exp\left(-y\|x_i - x_j\|^2\right)$. Here y is a constant used in the radial basis function.

**Random forest (RF)**

RF belongs to the category of ensemble learning algorithms, having been proposed by Ho in (Ho, 1998). As a base learner of the ensemble, RF uses decision trees. The idea of ensemble learning is that a single predictor is not sufficient for predicting the desired value of test data. The reason being that, based on sample data, a single predictor is not able to distinguish between noise and patterns. RF constructs numerous independent regression trees, a bootstrap sample of the training data is chosen at each regression tree. Therefore, the regression tree continues to grow until it reaches the largest possible size. Whereas, final prediction values a weighted average from predicting all regression trees (Breiman 2001).

**Deep learning (DL)**

Deep Learning extends classical ML by adding more "depth" (complexity) into the model, as well as transforming the data using various functions that create data representations in a hierarchical way, through several levels of abstraction. A strong advantage of DL is feature learning, i.e., automatic feature extraction from raw data, with features in higher levels of the hierarchy being formed through composition of lower level features. DL can solve complex problems particularly well and fast, due to the more complex models used, which also allow massive parallelization. These complex models employed in DL can increase classification accuracy, or reduce error in regression problems, provided there are adequately large datasets available describing the problem. DL includes different components, such as convolutions, pooling layers, fully connected layers, gates, memory cells, activation functions, encoding/decoding schemes, depending on the network architecture used, e.g., Convolutional Neural Networks, Recurrent Neural Networks, Unsupervised Networks (Kamilaris et al., 2018).

**Long short-term memories (LSTM)**

The LSTM model was initial introduced in (Hochreiter and Schmidhuber, 1997) with the objective of modeling long term dependencies and determining the optimal time lag for time series problems. A LSTM network is composed of one input layer, one recurrent hidden layer, and one output layer. The basic unit in the hidden layer is the memory block, containing memory cells with self-connections memorizing the temporal state and a pair of adaptive, multiplicative gating units controlling information flow in the block. The memory cell is primarily a recurrently self-connected linear unit, called Constant Error Carousel (CEC), and the cell state is represented by the activation of the CEC. The multiplicative gates learn when to open and close. By keeping the network error constant, the vanishing gradient problem can be solved in LSTM. Moreover, a forget gate is added to the memory cell preventing the gradient from exploding when learning long time series. The operation and structure of LSTM can be described as follows:

$$i_t = \sigma(w_i x_t + U_i m_{t-1} + b_i)$$

$$s_t = \tanh(w_s x_t + U_s m_{t-1} + b_s)$$

$$f_t = (w_s x_t + U_s m_{t-1} + b_s)$$

$$f_t = \sigma(w_f x_t + U_f m_{t-1} + b_f)$$

$$c_t = c_{t-1} \circ f_t + s_t \circ i_t$$

$$m_t = s_t \circ o_t \tag{1}$$

where $i_t$, $i_t$ and $y_t$ are denoted as input gate, forget gate and output gate at time t respectively, and $m_t$ and $c_t$ represent the hidden state and cell state of the memory cell at time t.

**Microclimatic measurements**

In our first experiment, the DL and ML models were applied to Ficus plants (Ficus benjammina) data collected from four cultivation tables in a $90m^2$ greenhouse compartment of the Ornamental Plant Research Centre (PCS) in Destelbergen, Belgium. Plant density was approximately 15 pots per $m^2$, where every pot contained 3 cuttings. Greenhouse microclimate was set by controlling the window openings, a thermal screen, an air heating system, assimilation light and a CO2 adding system. Plants were irrigated with an automatic flood irrigation system, controlled by time and radiation sum. Set points for microclimate and irrigation control were similar to the ones used in commercial greenhouses. The microclimate of the greenhouse was continuously monitored. Photosynthetic active radiation (PAR) and CO2 concentration were measured with a LI-190 Quantum Sensor (LI-COR, Lincoln, Nebraska, USA) and a carbon dioxide probe (Vaisala CARBOCAP GMP343, Vantaa, Finland), respectively. Temperature and relative humidity were measured with a temperature and relative humidity probe (Campbell Scientific CS215, Logan, UT, USA), which was installed in a ventilated radiation shield. Stem diameter was continuously monitored on these plants with a linear variable displacement transducer (LVDT, Solartron, Bognor Regis, UK) sensor. The hourly variation rate of stem diameter (mm $d^{-1}$) was calculated as the difference between the current stem diameter and the stem diameter recorded on hour early for a given time point.

In the second experiment, the DL and ML models have been trained with data collected from a UK Greenhouse farm, including both environmental (CO2, humidity, radiation, outside temperature, inside temperature), as well as, yield actual measurements. The environmental data were collected on an hourly basis, while the yield on a weekly basis. To deal with these data characteristics, we performed data augmentation, through interpolation of weekly data, resulting in daily data measurements. We also performed averaging of the hourly environmental data, so as to achieve similar daily representations.

In both experiments, the data were split into training, testing and validation datasets. 60% of the recorded data were assigned to the training set, 15% to the validation set and 25% to the test set.

**Prediction evaluation**

The Mean Absolute Error (MAE), Root Mean Squared Error (RMSE) and Mean Squared Error (MSE) have been used to evaluate the performance of these prediction models. Formulas of these evaluation measures are shown in the following Equations:

$$\text{MSE} = \frac{1}{n}\sum_{t=1}^{n}\left(\frac{A_t - F_t}{A_t}\right)^2 \tag{2}$$

$$\text{MAE} = \frac{1}{n}\sum_{t=1}^{n}\frac{|A_t - F_t|}{|A_t|} \tag{3}$$

$$\text{RMSE} = \sqrt{\frac{1}{n}\sum_{t=1}^{n}\left(\frac{A_t - F_t}{A_t}\right)^2} \tag{4}$$

where $A_t$ is the actual value and $F_t$ is the predicted value.

**RESULTS AND DISCUSSION**

We have developed and tested DL (LSTM), SVR and RFR prediction models to predict plant yield and growth in greenhouse environments for: a) ficus growth prediction based on the SDV indicator, b) tomato yield prediction. A commonly used method, grid search, was utilized to determine the parameters of each model. The parameters gamma and C were of importance for the SVR model design. The number of trees in addition to max depth of the tree were of importance in the RF model design. The number and size of hidden layers were of importance for the DL LSTM model design.

The implemented approach involved three steps:

- Data preprocessing and data cleaning.
- Data splitting into training, validation and test datasets.
- DL/LSTM, SVR, and RF model design and use to generate one step ahead prediction.

The obtained results clearly show that the DL/LSTM model outperforms he SVR and RF ones, in both experiments. Table 1 shows the obtained accuracy, in terms of MSE, RMSE and MAE, when each of the (trained) three models is applied to the test datasets, in both experiments.

Table 1. Performance of the DL/LSTM model compared to those of SVR and RF models for plant yield and growth prediction.

| Datasets | Tomato Yield | | | Ficus Growth(SDV) | | |
|---|---|---|---|---|---|---|
| Models | SVR | RF | LSTM | SVR | RF | LSTM |
| MSE | 0.015 | 0.040 | **0.002** | 0.006 | 0.006 | **0.001** |
| RMSE | 0.125 | 0.200 | **0.047** | 0.073 | 0.062 | **0.042** |
| MAE | 0.087 | 0.192 | **0.03** | 0.070 | 0.063 | **0.030** |

Figure 1 illustrates the performance of prediction models (RF, SVR and LSTM). It is clear that LSTM model was able to successfully predict Ficus growth (SVD) and outperformed RF and SVR. Figure 2 shows how LSTM model followed the trend of the actual yield value and showed the ability to store a better representation of the temporal nature of the given data and, to this end, generalize better than RF and SVR models.

**CONCLUSIONS**

The paper developed a DL approach using LSTM for Ficus growth (represented by the SDV) and tomato yield prediction, achieving high prediction accuracy in both problems. Experimental results were presented that show that the DL technique (using a LSTM model) outperformed other traditional ML techniques, such as SVR and RF, in terms of MSE, RMSE and MAE error criteria. Hence, the main aim of our project is to develop DL methodologies to predict plants growth and yield in greenhouse environment. Future studies looking at the continuity of : a) greatly increase the number of collected data that are used for training the proposed DL methods; b) extending the DL method so as to perform multi-step (at a weekly, or a multiple of weeks basis) prediction of growth and yield in a large variety of greenhouse, in the UK and Europe.

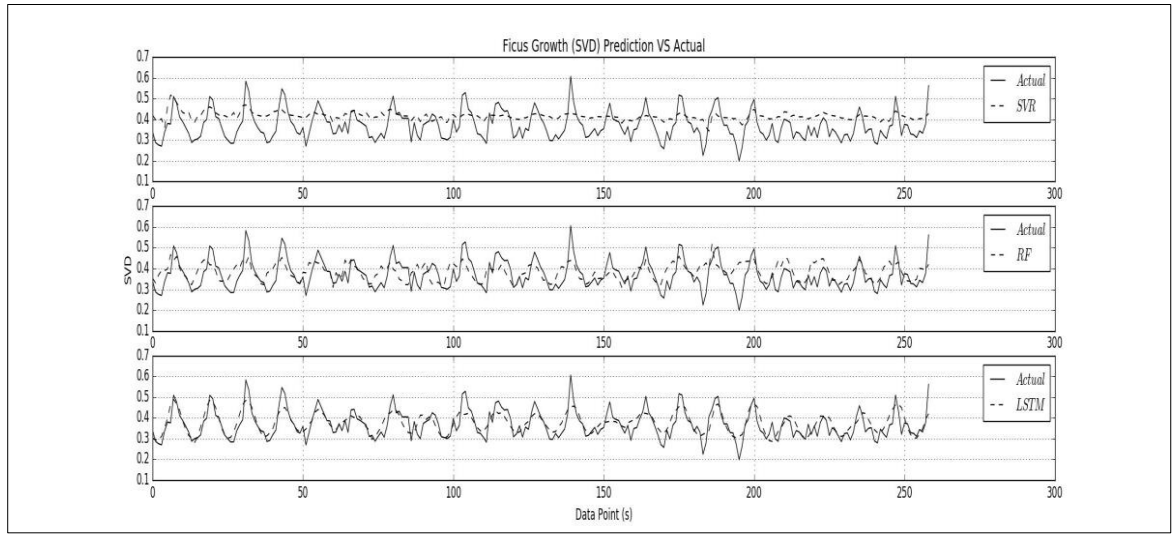

Figure 1. Testing results and performance comparison of Ficus growth (SVD) predictions.

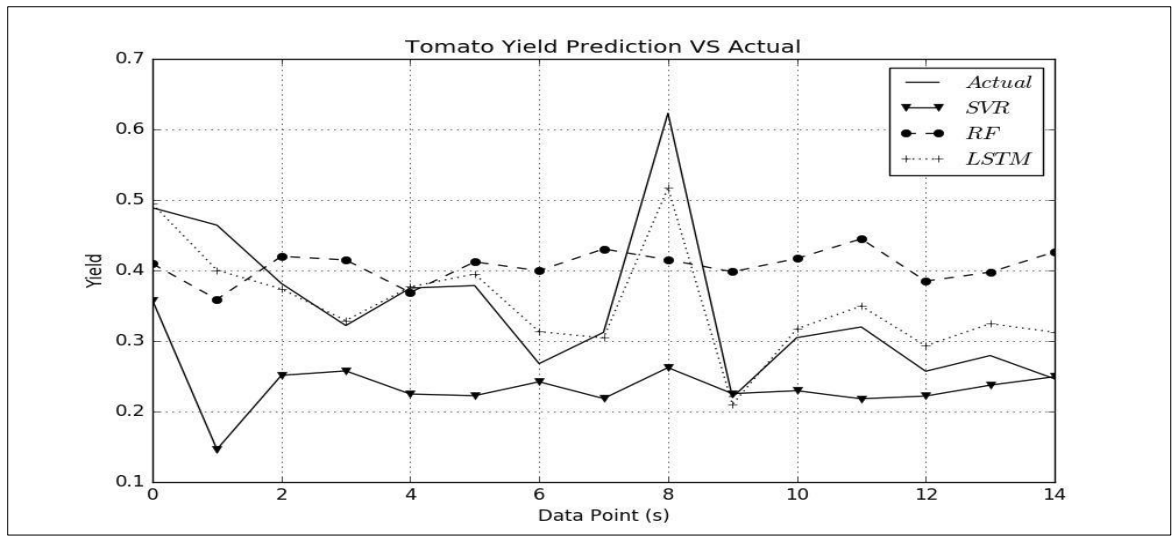

Figure 2. Testing results and performance comparison of Tomato Yield predictions.

**ACKNOWLEDGEMENTS**

This work is part of EU Interreg SMARTGREEN project (2017-2021). We would like to thank all the growers (UK & EU), for providing the data; Their valuable feedback, suggestions and comments are highly appreciated to increase the overall quality of this work.